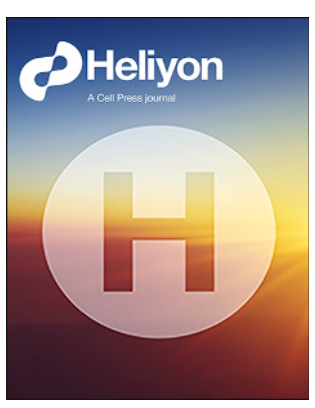

Research article

# A systematic review of machine learning techniques to address diagnosis and treatment of autism: challenges and opportunities

Rafael Muñoz-Terol [a,*], Jesús Peral [a], Sandra Amador [b], David Gil [c]

[a] Lucentia Research Group, Department of Software and Computing Systems, University of Alicante, 03690 Alicante, Spain
[b] U.I. for Computer Research, Alicante, Spain
[c] Department of Computer Science Technology and Computation, University of Alicante, 03690 Alicante, Spain

ABSTRACT

Autism spectrum disorder (ASD) is a developmental disability characterized by challenges in social interaction and communication. As the causes of ASD remain unclear, identifying relevant features and hidden correlations is crucial for early diagnosis. This systematic review evaluates 55 studies from 2017 to 2023 on the application of machine learning (ML) techniques to ASD. The primary objective is to examine recent ML applications in ASD research, identifying trends, techniques, and datasets that enhance diagnosis and treatment. Supervised learning methods dominate, as they align well with ASD diagnostic needs; however, the role of deep learning is expanding with greater data availability. Emerging techniques based on hybrid methods, where unsupervised, deep learning, and fuzzy logic could be included, will be interesting to observe in the future. The review highlights key challenges and opportunities, particularly the need for models that can integrate complex data —such as genetic and clinical information— to improve diagnostic accuracy and treatment outcomes. Additionally, incorporating innovative data sources, like wearable devices and biometric sensors, could enable continuous and non-intrusive monitoring, providing a more holistic understanding of ASD. Findings emphasize that addressing current challenges requires interdisciplinary collaboration and expanded datasets tailored to ASD. Future ML models will benefit from broader multimodal data integration, enabling researchers to more comprehensively address the complexities of ASD.

## 1. Introduction

Over the past few years, research on autism spectrum disorder (ASD) has been one of the most important topics in computational psychiatry. ASD is a developmental disability characterised by impaired social communication and social interactions, in addition to restricted and constant patterns of conduct (Diagnostic and Statistical Manual of Mental Disorders[1], DSM-5). Therefore, ASD is a divergent disability related to the appearance of symptoms and severity, risk factors, study, and treatment responses. Regarding computational psychiatry and ASD research frameworks, various machine learning (ML) approaches have been used [1–3], which have demonstrated better results than the knowledge-based approaches. Therefore, in these children psychiatry-applied framework, the best doctors' decisions will enhance the well-being of patients with ASD. This fact implies that the application of ML approaches to the ASD problem is a research trend in detection and management tasks. In this way, ML algorithms are traditionally classified into two different categories according to their learning methods: unsupervised and supervised learning. Thus, as noted by Lloyd et al. [4], in supervised learning, the machine assumes the role from a set of training examples, whereas in unsupervised learning, the machine attempts to detect hidden structures in unlabelled data. Moreover, there is a third category of ML classification algorithms for studying the ASD problem, namely, hybrid algorithms that combine ML techniques and knowledge-based resources.

* Corresponding author.
*E-mail addresses:* rafamt@dlsi.ua.es (R. Muñoz-Terol), jperal@dlsi.ua.es (J. Peral), saandra.amador@gmail.com (S. Amador), david.gil@ua.es (D. Gil).

[1] https://www.psychiatry.org/Psychiatrists/Practice/DSM/Updates-to-DSM/Coding-Updates/2021-Coding-Updates (visited on November 20, 2025).

Various ML-based approaches have been implemented in ASD detection and observation frameworks. Crippa et al. [5] conducted an evidence-of-concept study to determine whether an easy higher-limb gesture could be used in categorising low-functioning infants (aged 2–4 years) with ASD by developing a supervised ML method that correctly identifies 15 kindergarten infants with ASD from 15 commonly developing infants through the kinematic analysis of a simple reach-to-drop task. In addition, Duda et al. [6] trained and tested three pairs of ML models on the full 65-element Social Responsiveness Scale result sheets from a set of individuals with either ASD or attention deficit hyperactivity disorder (ADHD), whose scores justified the suitability of this method for distinguishing ASD and ADHD with high accuracy. Bone et al. [7] utilized ML to infer ASD instrument algorithms and enhance commonly used ASD screening and diagnostic tools by applying support vector machine (SVM) approaches. Usta et al. [8] tested the performance of four ML techniques: naive Bayes, generalized linear model, logistic regression, and decision trees (DTs). They concluded that these ML models indicate that several others are more remarkable in terms of predictive information and management of the treatment of children with ASD. Rudovic et al. [9] used the latest approaches in deep learning to propose a personalised ML framework for the automatic discrimination of the emotional behaviours of children. Furthermore, it includes the arrangement during robot-assisted autism treatment exhibiting the viability of robot perception of sweetie and commitment in infants with autism and has inferences in the design process of future autism treatments. Thabtah and Peebles [10] proposed a new ML method that in addition distinguishes autistic features of situations and checkpoints and provides users with knowledge rules that can be utilized by domain professionals to recognize the purposes after the classification. Thus, empirical scores from three data sets show that rule-based machine learning provides classifiers with better predictive accuracy, harmonic mean, specificity, and sensitivity than other ML approaches. Tariq et al. [11] theorised that ML tests performed on home videos could rate an assay without compromising accuracy. Consequently, they tested item-level records using a pair of standard analytical tools to develop ML classifiers that enhanced spareness, explainability, and accuracy. They eventually checked whether features from the improved models could be considered by blinded amateur raters from 3-min home videos of infants with and without ASD for rapid and correct ML autistic classification. Liu et al. [12] applied an ML method to detect an eye motion dataset from a face recognition task to categorise infants with and without ASD. Thus, they analysed the performance of the model by considering its sensitivity, accuracy, and specificity in categorising ASD; the scores confirmed the utility of the ML algorithm that manages the face-scanning patterns for identifying infants with ASD.

ML hybrid algorithms combine multiple ML techniques or models to develop a new approach that leverages the strengths of each individual component. These algorithms aim to enhance the performance, improve accuracy, and tackle specific challenges that a single technique cannot adequately resolve. In recent years, the research community has explored different approaches for developing hybrid algorithms. These methods include ensemble methods [13], feature selection/extraction combinations [14], and model stacking [15]. Ensemble methods involve a combination of multiple base models, such as DTs, neural networks, and SVMs in the prediction process. In contrast, feature selection/extraction combinations mix different subsets of features or data representations to improve the final prediction or decision. Finally, model stacking trains multiple models on the same dataset and combines their predictions using another model, often referred to as a meta-model or blender.

Hybrid-based approaches have also been applied to the ASD problem. Alharbi et al. [16] presented an expert system to evaluate autism under unpredictability that used trusted knowledge-based inference methods with the evidential reasoning technique, where the knowledge base of the system was obtained by utilizing real data from affected people, and by considering doctors' opinions. Alam et al. [17] proposed an Internet-of-Things belief knowledge base-based hybrid system. This intelligent approach can instinctively identify the signs and symptom data of different autistic infants in real-time and categorised them. The belief-rule-base subsystem included knowledge description variables to accomplish this goal. Thus, infants with autism were categorised based on the signs and symptoms identified by pervasive sensing components. Mugzach et al. [18] developed an ontology that allowed data integration and reasoning with patient data to categorise subjects, and deduce new rules based on ASD and related neurodevelopmental disorders based on this categorization. Gong et al. [19] demonstrated a method to prognosticate autism propensity genes, where genes were first extracted from biomedical literature, and some autism liability genes were then perceived as kernels by the superior understanding. Persisting genes were predicted by developing alliance rules between the kernels and candidates.

We can summarise that hybrid learning techniques developed in recent years have shown great potential to improve diagnostic accuracy and personalize treatments in the context of ASD. Unlike traditional methods, hybrid models combine multiple machine learning techniques, such as integrating neural networks with rule-based systems or optimization algorithms. This approach provides greater flexibility and adaptability in the classification and analysis of complex data, enabling a more detailed understanding of patterns associated with autism. At the same time, deep learning has gained traction due to its ability to recognize highly complex patterns in large volumes of multimodal data. Advanced neural network models and deep learning architectures, such as convolutional neural networks (CNNs) or recurrent neural networks (RNNs), could be integrated into future studies to improve the detection of the early ASD markers. Implementing transfer learning techniques, for example, would allow models to leverage knowledge from similar domains to enhance accuracy in limited ASD datasets. Additionally, explainable AI (XAI) holds the potential to provide interpretability in these models, helping medical professionals understand and justify algorithm-based decisions. These developments present a promising path for future research, combining the precision of deep learning with the clarity and adaptability needed in a medical context.

As can be observed from all the studies presented, since the 2010s a series of works have laid the foundations for the application of ML techniques for the detection, diagnosis, and treatment of ASD [5–7,12,13,16,20]. However, after analysing the previous state-of-the-art studies, we believe that a current survey that reviews ML techniques, their suitability and performance, as well as the creation and accessibility of datasets in autism is needed. Consequently, this systematic review aims to summarise the current ML-based approaches that have been applied to the ASD problem in order to address the research questions described in Table 1. We have focused on the last seven years, from 2017 to 2023, to provide an overview of the most modern trends in this area. First, the exploration strategies utilized to retrieve the relevant literature published over the last seven years are outlined. The various studies that describe the predominant techniques, venues, databases, and various performance metrics are also explained. Subsequently, the limitations, challenges, and metrics of all the analysed ML algorithms are described in the Discussion section. Finally, the conclusions and future research directions are presented.

## 2. Methods and materials

The next two subsections describe the search strategies, the research questions and the criteria for choosing the studies included in this review.

### 2.1. Search strategy and research questions

In this review, a search strategy was developed to retrieve studies related to ASD. The search task was performed using major database collections, including ScienceDirect, Scopus, and Multidisciplinary Digital Publishing Institute (MDPI), with the keywords “autism spectrum disorder”, “autism”, “ASD”, and “machine learning”. The review questions (RQs) used in this review are classified into eight items (Table 1).

### 2.2. Study selection

This section outlines the methodology employed to systematically select, review, and classify studies on the application of ML techniques to ASD. To retrieve relevant research works, manuscripts were selected based on well-defined inclusion and exclusion criteria:

1. Inclusion Criteria:
   - Studies published between 2017 and 2023, capturing recent advancements and trends in ML applications to autism research, reflecting the latest technological innovations and developments in the field.
   - Manuscripts written in English to ensure consistency in interpretation and to align with the scope of our review.
   - Research works that specifically addressed ASD, autism, and the application of ML methods, with a focus on diagnosis, intervention, or behavioural analysis.
2. Exclusion Criteria:
   - Manuscripts that did not directly address ASD or apply ML to ASD-specific challenges.
   - Studies comprised of abstracts, conference proceedings, review articles, tutorials, or non-peer-reviewed materials.
   - Manuscripts focusing on ML applications outside of the ASD context, even if they used similar techniques in other domains.

The process began by identifying studies through systematic keyword searches in ScienceDirect, Scopus, and MDPI databases, resulting in 163 initial publications (83 from ScienceDirect and 80 from Scopus and MDPI) that met the basic inclusion criteria. After duplicate removal, 92 records remained for initial screening. Records that were abstracts, conference proceedings, review articles, tutorials, non-English publications, or unrelated to the topic (n = 37) were then excluded, resulting in 55 final studies.

**Table 1**
Research questions that have been applied to the research studies.

| ID | Question | Rationale/Motivation |
|---|---|---|
| RQ1 | Where and when were the previous studies published? | To determine the quality of scientific contributions. |
| RQ2 | Which countries are the authors of the contributions based in? | To identify the geographic distribution of contributions. |
| RQ3 | Which type of ML approach is the most frequently used? | To identify the most frequently investigated types of approaches (including combined methods) proposed in literature. |
| RQ4 | Which ML techniques are the most frequently used? | To identify the most frequently adopted techniques to develop ensemble approaches in ASD diagnosis in literature. |
| RQ5 | Which data types are used most frequently in the selected studies? | To identify the most frequently investigated data types. |
| RQ6 | How many patients do the selected studies examine? | To determine the number of patients used for ASD studies. |
| RQ7 | Do they normally develop their own databases in their studies? If not, which public databases are the most frequently used? | To identify the most commonly used public databases and determine the number of studies that created their own databases. |
| RQ8 | Are ML techniques a good choice for managing ASD? | To analyse the good performance of ML techniques according to the different features proposed in the literature. |

Each retrieved manuscript was independently reviewed by at least two researchers experienced in computation and data science (ML). The abstracts, methods, and results were carefully analysed to ensure that only studies of high relevance and methodological soundness were included. Fig. 1 presents the PRISMA flowchart of our selection process [21,22], with all inclusion and exclusion steps outlined. Utilizing the PRISMA methodology enhances review quality and consistency, which is particularly beneficial for systematic reviews in medical and clinical research. This methodology ensures that our review focuses exclusively on high-quality research addressing ML applications in the ASD field, capturing significant trends, techniques, and datasets relevant to current and future advancements.

## 3. Results

The following eight subsections describe the study conducted to answer each of the research questions regarding the review framework.

### 3.1. RQ1: Where and when were the previous studies published?

These studies were published as high-quality contributions in international journals over the past seven years (between 2017 and 2023). All these international journals are ranked in the Journal Citation Report according to their impact factors that place them in one of the four quartiles: Q1, Q2, Q3 or Q4. Moreover, according to the journal-publishing countries listed, various journals from the Netherlands published the largest number of the research works (27,27 %), as shown in Table 2.

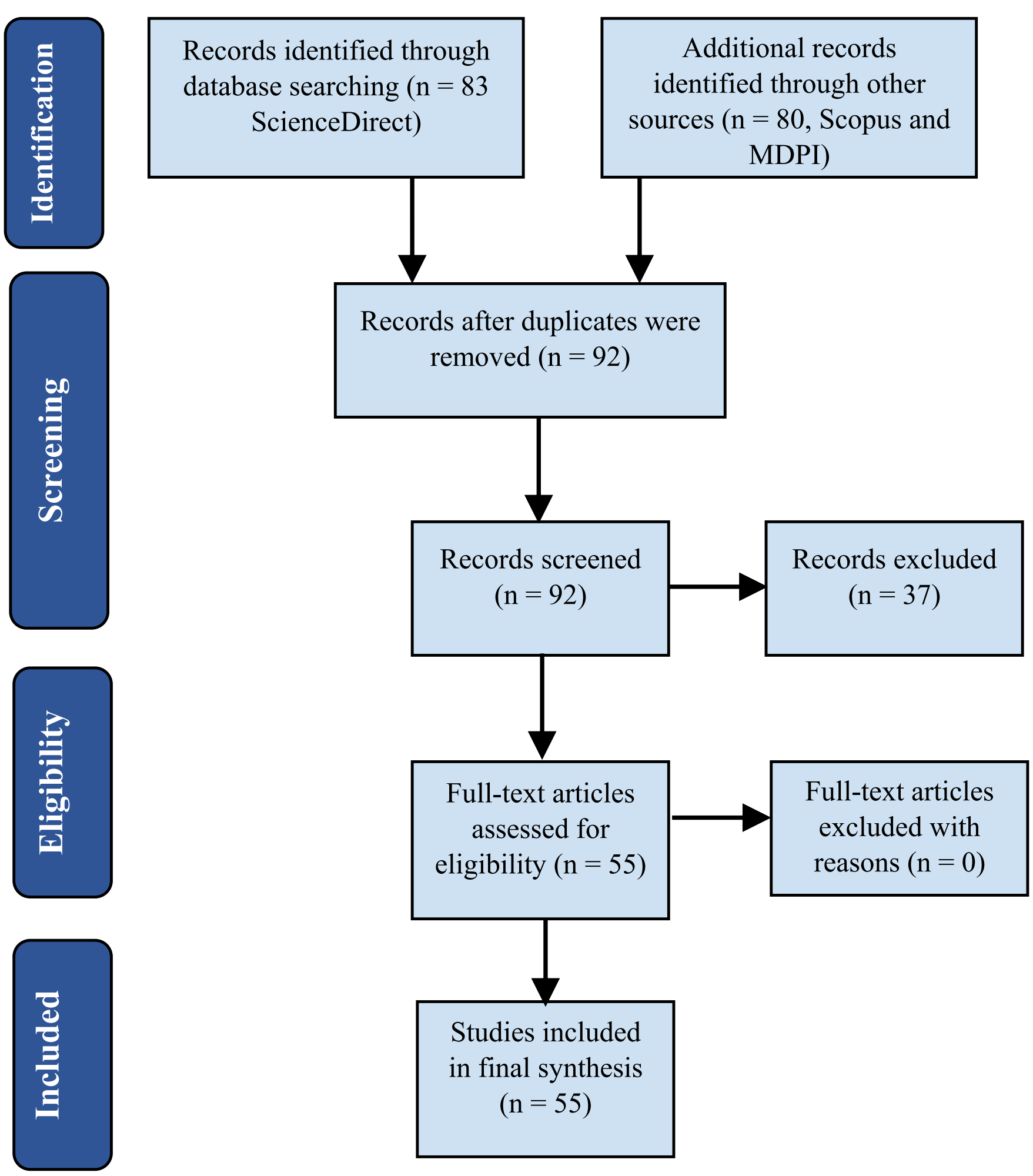

**Fig. 1.** PRISMA flowchart that shows the process applied to the manuscripts considered in this study.

**Table 2**
Countries contributing to journal publications detailing where research studies related to RQ1 were published.

| Country | Studies | % Studies |
|---|---|---|
| Netherlands | [23–37] | 27,27 % |
| USA | [38–49] | 21,81 % |
| UK | [10,50–59] | 20 % |
| Ireland | [60–65] | 10,91 % |
| Switzerland | [66–71] | 10,91 % |
| India | [72] | 1,82 % |
| Iran | [73] | 1,82 % |
| Sweden | [74] | 1,82 % |
| Poland | [75] | 1,82 % |
| Egypt | [76] | 1,82 % |

### *3.2. RQ2: Which countries are the authors of the contributions based in?*

Table 3 lists the number of contributions from each country considering two aspects. First, the column labelled as “corresponding_author_contributions” ranks the countries according to the country to which the corresponding author's institution belongs. Second, the column labelled as “remaining_author_contributions” ranks the countries according to the country to which the remaining authors' institutions belong.

### *3.3. RQ3: Which type of ML approach is the most frequently used?*

Table 4 presents the different ML approaches used in all the studies.

Interestingly, supervised learning approaches were used most frequently and obtained the highest percentage in detection tasks for individuals on the autism spectrum (Fig. 2). This was expected because the classification was known and supervised. However, it is important to consider the increase in the number of hybrid algorithms, which in many cases complement supervised learning via clustering techniques through a separate phase-by-phase process.

As can be seen, most of the work relies on supervised methods since classification (autistic diagnosis) is a clear objective. The strengths of these techniques are thereby that they are effective for classification and prediction when labelled data are available, allowing for greater accuracy in pattern identification. Therefore, they are useful in the early detection of ASD through the identification of specific markers and the classification of symptoms associated with ASD. However, even though this is a strength, it can become a limitation when the dataset is large or when some records are not labelled where unsupervised techniques may be more appropriate. Their weaknesses are therefore the dependence on the quality and quantity of available labelled data, which may be limited in ASD clinical settings, as well as the risk of overfitting if training data do not adequately represent the diversity of the autistic spectrum. Alternatively, unsupervised techniques have the strengths of allowing the discovery of underlying patterns in the data without requiring prior labelling, which is useful for comprehensive data exploration, facilitating the identification of unique developmental profiles within the autistic spectrum. Weaknesses are that interpretation of results can be challenging due to lack of labels, which could limit direct clinical application. They are sensitive to data quality and the presence of noise, which may affect the accuracy of results. They may have applicability in helping to identify subgroups within the autism spectrum, which can guide personalised interventions and therapeutic approaches. In many cases, combining both methods (supervised and unsupervised) can lead to hybrid techniques that leverage the strengths of both strategies.

The algorithms listed as “Others” refer to statistical techniques or ensembles of comparative algorithms. Algorithms that are not included in the other categories are classified in this category.

### *3.4. RQ4: Which ML techniques are the most frequently used?*

Table 5 lists the ML algorithms used in these studies and their corresponding year intervals (2017–2018, 2019–2020, and 2021–2023).

Fig. 3 shows that “classical” algorithms were used most frequently. However, it should be noted that in the coming years, new algorithms utilizing cutting-edge methodologies, such as deep learning, could become the most frequently used.

The algorithms listed as “Others” include those that are not categorised in the other categories.

### *3.5. RQ5: Which data types are used most frequently in the selected studies?*

Table 6 lists the different data types that were used in the selected studies.

Fig. 4 shows that brain data are the most frequently utilized data type in these studies (45 %), followed by clinical data (25 %) and eye tracking (18 %). In contrast, the ballistocardiogram (BCG), prosody, and phenotype were used only once. Moreover, there are three special cases in which the same study used different data types: specifically, the cases that use brain data and eye tracking [53], clinical data and eye tracking [29], and a combination of brain data, eye tracking, and facial recognition [76].

**Table 3**
Countries of the corresponding and remaining authors' institutions regarding RQ2.

| Country | corresponding_author_contributions | remaining_author_contributions |
|---|---|---|
| USA | [25,29,30,33,38,41–43,46,47,51,61,73] | [23–25,29,30,33,38,39,41–43,46,47,51,61,66,73] |
| China | [32,39,40,50,52–54,60,64,74,76] | [24,32,39,40,50,52–54,60,64,65,74,76] |
| India | [24,28,49,70] | [24,28,49,66,68,70,72] |
| Spain | [31,36,67] | [31,36,44,67] |
| Italy | [26,56,57] | [26,56,57] |
| Bangladesh | [48,71,72] | [48,71,72] |
| United Kingdom | [10,62] | [10,27,34,35,45,68] |
| Saudi Arabia | [68,75] | [62,71,75] |
| Germany | [59,69] | [59,69] |
| Brazil | [35,45] | [35,45] |
| Turkey | [27,34] | [74] |
| Australia | [37] | [37,43,48,71] |
| Canada | [23] | [23,44,46] |
| Egypt | [55] | [41,55] |
| Finland | [44] | [23] |
| Portugal | [63] | [63] |
| Singapore | [66] | [54] |
| Iraq | [58] | [58] |
| United Arab Emirates | – | [58,62,70] |
| Malaysia | – | [58,68] |
| Austria | – | [59] |
| Iran | – | [67] |
| Japan | – | [66] |
| Jordan | – | [66] |
| Pakistan | – | [48] |
| Sweden | – | [59] |
| Taiwan | – | [66] |
| Tunisia | – | [34] |

**Table 4**
ML method types used in the research studies regarding RQ3.

| ML algorithms | Studies |
|---|---|
| Supervised Learning | [10,23,25,27,29,32–34,38–40,42,44–47,49–51,53,55–57,59,63,64,66,67,69,71–75] |
| Unsupervised Learning | [35,60,61] |
| Hybrid | [24,26,28,30,31,43,48,52,58,68,70,76] |
| Others | [36,37,41,54,62] |

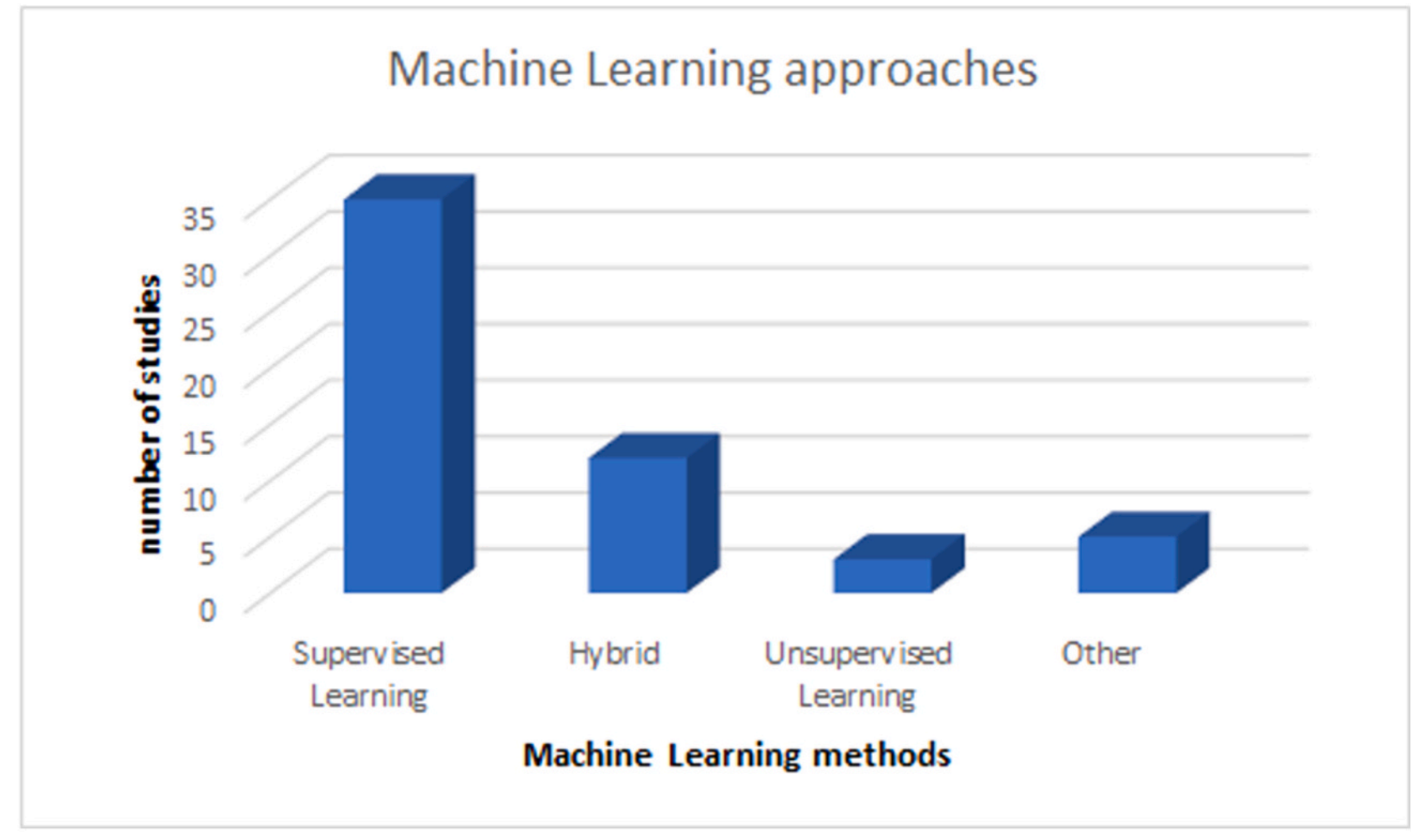

**Fig. 2.** Machine learning approaches used in the research studies regarding RQ3.

**Table 5**
Classification of ML Algorithms, subtypes, years interval, and its application in the research studies: extended information to RQ4.

| ML Algorithm type | Algorithm | Years interval and # of studies | Studies |
|---|---|---|---|
| Supervised Learning | SVM | 2017–2018 (6) | [35,43,44,46,56,63] |
| | | 2019–2020 (16) | [23–25,27,32–34,40,45,50–53,64,66,67] |
| | | 2021–2023 (9) | [37,38,47–49,58,65,70,72] |
| | Random Forest (RF) | 2017–2018 (4) | [35,42,56,57] |
| | | 2019–2020 (10) | [28–30,37,40,49,51,67,73,74] |
| | | 2021–2023 (5) | [48,58,59,69,72] |
| | MLP (Neural Network) | 2017–2018 (1) | [35] |
| | | 2019–2020 (5) | [25,31,40,64,67] |
| | | 2021–2023 (4) | [49,58,65,68] |
| | Decision Trees (DT) | 2017–2018 (1) | [62] |
| | | 2019–2020 (9) | [10,27–29,40,51,55,64,67] |
| | | 2021–2023 (6) | [48,49,58,65,69,70] |
| | k-nearest neighbours (KNN) | 2017–2018 (2) | [57,75] |
| | | 2019–2020 (5) | [27,28,33,51,66] |
| | | 2021–2023 (4) | [48,49,58,72] |
| | Bayes | 2017–2018 (1) | [57] |
| | | 2019–2020 (5) | [28,39,51,64,67] |
| | | 2021–2023 (5) | [37,48,49,58,76] |
| | Deep Learning, convolutional, autoencoders | 2017–2018 (1) | [35] |
| | | 2019–2020 (6) | [26,29,31,33,45,50] |
| | | 2021–2023 (1) | [70] |
| | Long short-term memory (LSTM) | 2019–2020 (1) | [32] |
| | Logistic Regression | 2017–2018 (3) | [44,56,57] |
| | | 2019–2020 (4) | [26–28,64] |
| | | 2021–2023 (6) | [37,48,49,58,65,72] |
| | Boosting & Classifiers, Bagging | 2019–2020 (2) | [10,27] |
| | | 2021–2023 (5) | [37,48,49,58,65] |
| | Modified Grasshopper Optimization Algorithm (MGOA) | 2019–2020 (1) | [28] |
| | Feature selection/PCA | 2019–2020 (5) | [31,32,53,66,67] |
| | Probabilistic neural network | 2019–2020 (1) | [66] |
| | Linear Discriminant Analysis (LDA) | 2021–2023 (1) | [48] |
| Unsupervised Learning | k-means Clustering | 2017–2018 (1) | [61] |
| | | 2019–2020 (9) | [24,52,60] |
| Others | Voting, ensembles classification | 2019–2020 (1) | [54] |
| | Statistics | 2019–2020 (1) | [41] |
| | Kernel Extreme Learning Machine | 2021–2023 (1) | [36] |

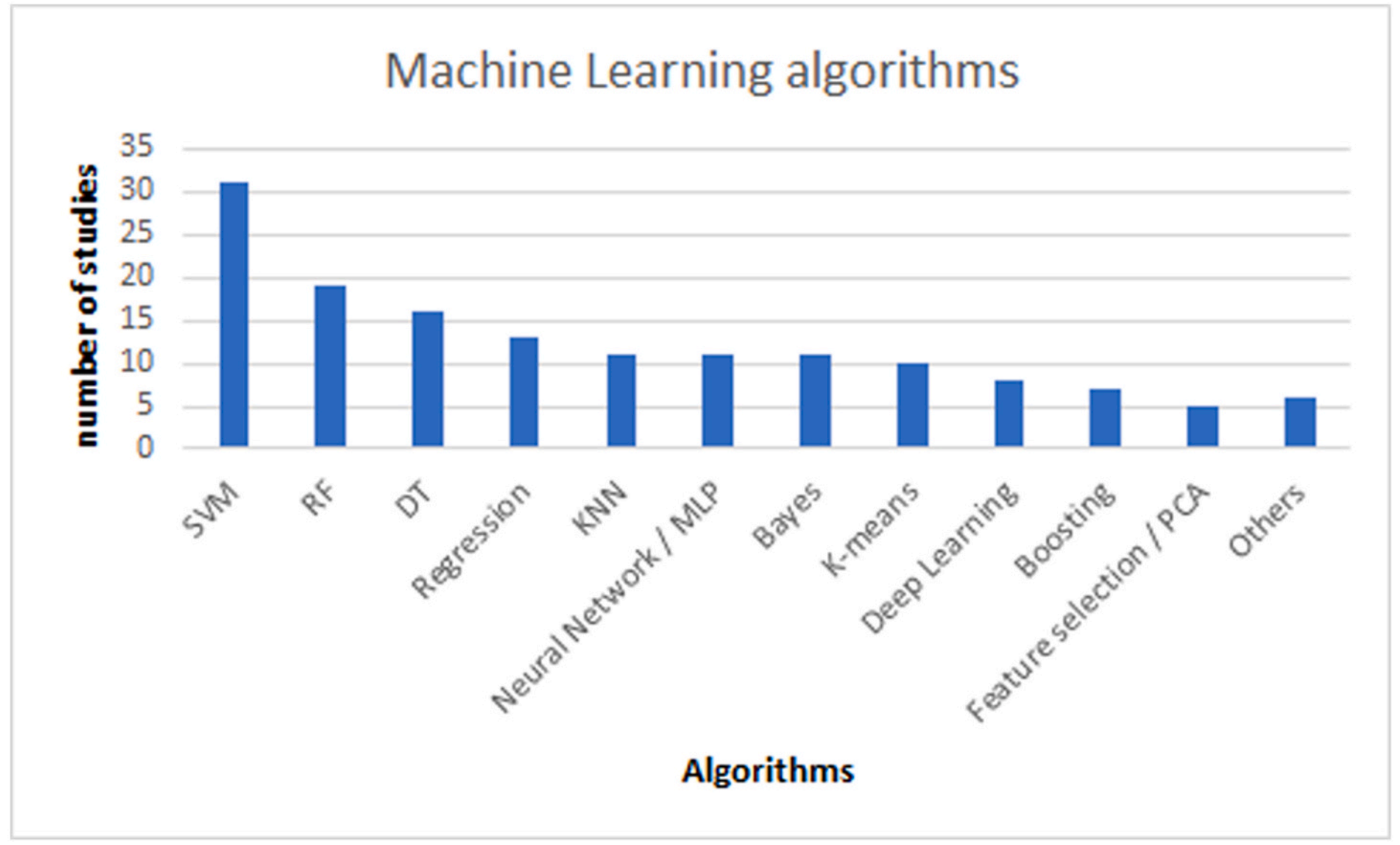

**Fig. 3.** ML algorithms used in the research studies in relation to RQ4.

**Table 6**
Data types used in the research studies with respect to RQ5.

| Data Type | Studies |
|---|---|
| Brain data | [23,24,26,27,30,31,34,35,38,39,42–46,53,54,56,57,60,63,64,66,75,76] |
| Clinical data | [10,28,29,37,48,49,55,58,59,62,65,67,71,72] |
| Eye tracking | [29,32,36,50–53,68,70,76] |
| Genome | [33,40,73,74] |
| BCG | [25] |
| Prosody | [47] |
| Phenotypes | [61] |
| Facial recognition | [41,69,76] |

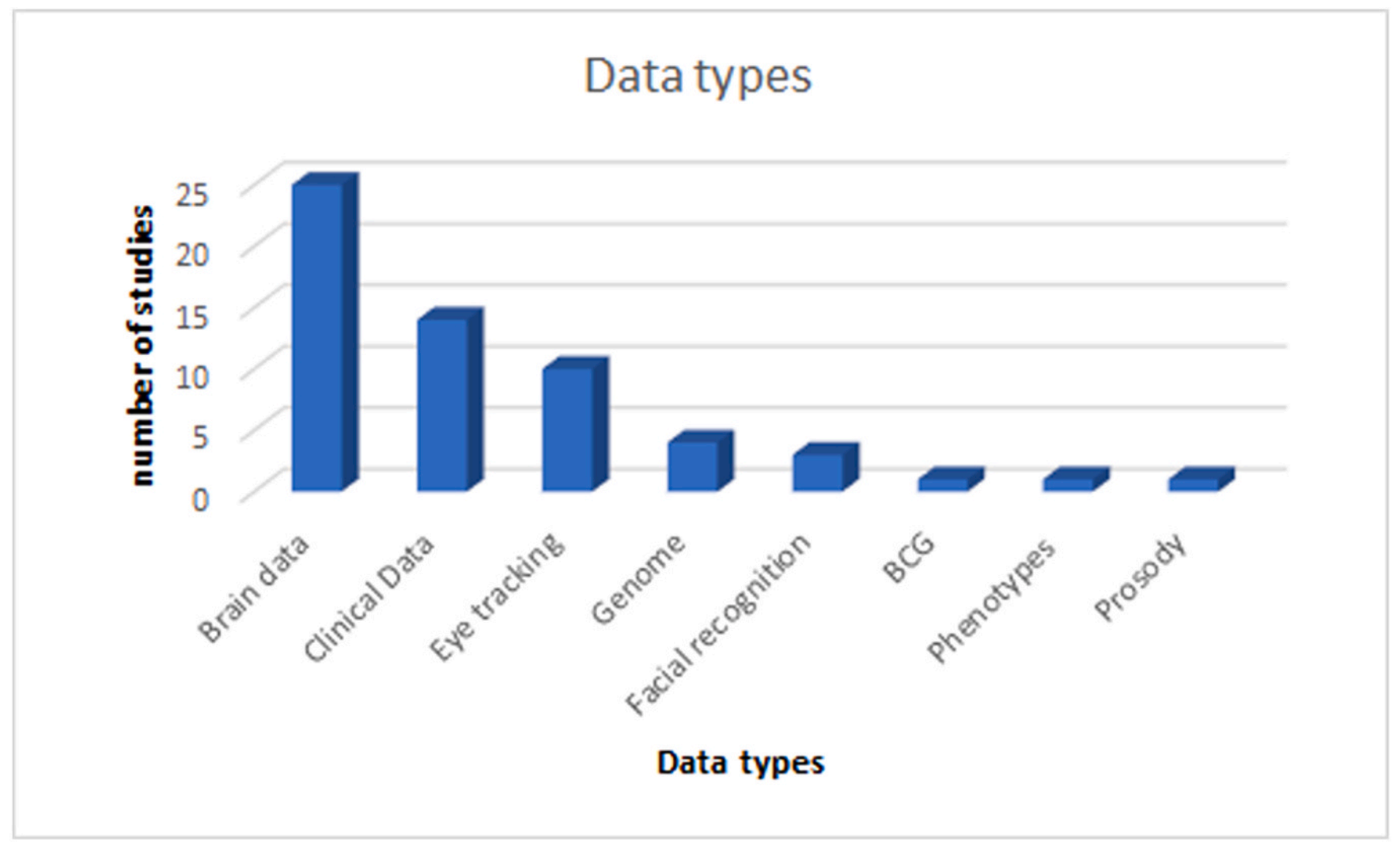

**Fig. 4.** Data types used in the research studies in relation to RQ5.

### 3.6. RQ6: How many patients do the selected studies examine?

Table 7 classifies studies according to the number of patients that were examined in these studies.

Fig. 5 shows the number of patients (sample sizes) selected by the studies to develop the performance tasks. In most of the studies, the sample sizes were 51–100 and 1001–2000 (20 % for each sample). However, for a few studies, the sample sizes were 201–300 (5.45 %).

After analysing the patient sampling, an attempt was made to correlate the ML algorithms with the datasets used to reduce biases and obtain greater generalisation and extrapolation using the ML algorithms. For this purpose, correlations between RQ4 and RQ6 were sought, and it can be concluded that, regarding deep knowledge, no specific algorithm produces a greater bias with the data. This conclusion should be verified in future studies with larger volumes of data.

### 3.7. RQ7: Do they normally develop their own databases in their studies? If not, which public databases are the most frequently used?

Table 8 lists the studies that developed their own databases and those that used public databases.

Considering the selected manuscripts, 42 % of them developed their own database, whereas 58 % used public databases (Fig. 6). Moreover [69,75], employed both strategies.

The research works shown in Fig. 7 used the following public databases: Autism Brain Imaging Data Exchange (ABIDE), Virulence Factor Database (VFDB), Sequence Read Archive database, Ambiente Di Ricerca Interdisciplinare Per L'Analisi Di Neuroimmagini Nell'Autismo (ARIANNA) database of the ARIANNA project, University of California Irvine (UCI) Machine Learning Repository, Karolinska Directed Emotional Faces (KDEF), Autism Genetic Resource Exchange (AGRE), Human Connectome Project (HCP), National Database for Autism Research (NDAR), NimStim Face Stimulus Set, Hartwell Autism Research, Technology Initiative (iHART), Figshare data repository, the ASD-Net database by the German research consortium, Kaggle, Autism Diagnostic Observation Schedule (ADOS-G/ADOS-2), and the National Insurance Institute (NII) dataset. The most frequently used public database was ABIDE (with a frequency of 31.42 %), followed by the UCI Machine Learning Repository. Table 9 lists the public databases used in the different studies.

**Table 7**
Number of examined patients in the research studies concerning RQ6.

| Number of patients | Studies |
|---|---|
| 0–50 | [23,25,33,41,51,57,60,63,64,75] |
| 51–100 | [36–38,46,50,53,66,68,70,74,76] |
| 101–200 | [27,30,34,42–44,47,52] |
| 201–300 | [29,32,40] |
| 301–1000 | [24,31,35,39,48,71] |
| 1001–2000 | [10,28,35,45,49,54,56,59,62,67,72] |
| >2000 | [26,55,58,61,65,69,73] |

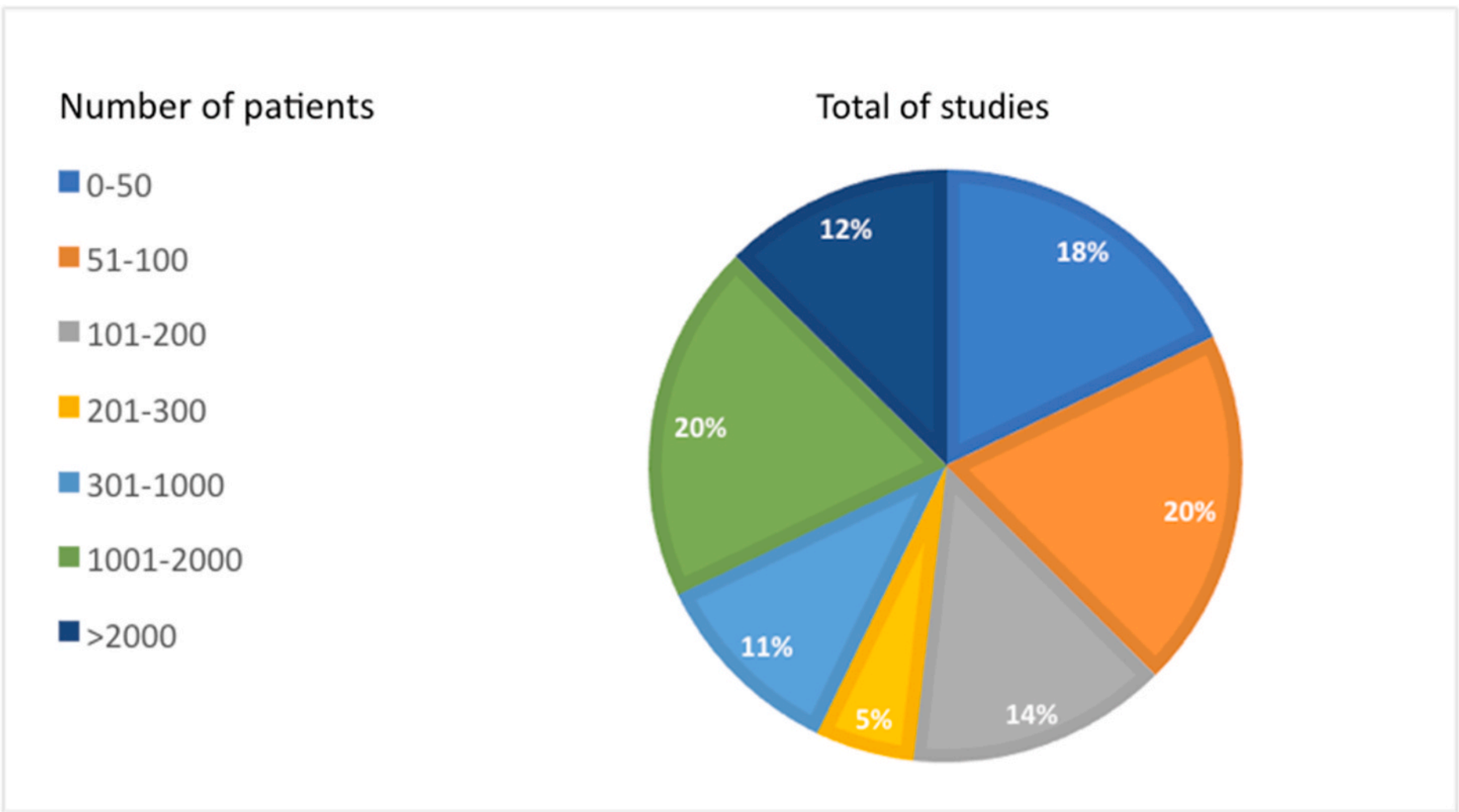

**Fig. 5.** Patient samples that were considered in the research studies related to RQ6.

**Table 8**
Studies that developed their own database or used a public database: extended information to RQ7.

| Database | Studies |
|---|---|
| Own database | [25,30,32,37,38,41,42,46,47,49–53,57,60,61,63–66,69,75,76] |
| Public database | [10,23,24,26–29,31,33–36,39,40,43–45,48,54–56,58,59,62,67–75] |

### *3.8. RQ8: Are ML techniques a good choice for managing ASD?*

The performance of the set of studies that applied ML techniques was analysed by considering the following five features: best ML method, accuracy (%), sensitivity (%), specificity (%) and area under the curve (AUC) (Fig. 8). Therefore, within the scope of the best ML method, any given ML method could be better because of the great diversity of ML methods that were highlighted as the best in the different environments considered by these studies. Thus, various supervised and unsupervised ML methods were adjudged to be the best methods for this set of studies. Moreover, a subset of these studies introduced deep learning methods as a novelty in this ASD research. Considering the accuracy (%), several research works have an accuracy of more than 90 %, indicating that ML-based approaches were the best for the ASD problem. Consequently, the sensitivity (%) scores obtained by these ML methods were as good as the previous accuracy (%) measures, including a small subset with a score of 100 %. It is well known that when a wide spectrum of approaches have different parameters, sensitivity analyses for model calibration must be performed to demonstrate their influences on the results. Thus, Asheghi et al. [77] demonstrated that parameters ranked using sensitivity analysis can be more reliable because they cover more uncertainties.

In addition, within the framework of the specificity (%) measure, the high scores obtained by most ML-based approaches (including scores of 100 %) also indicate that the application of ML methodology to the ASD research problem demonstrates the best performance in this computational biomedical domain. Finally, in terms of the AUC, the scores obtained by these ML methods also demonstrated their good performance for ASD research.

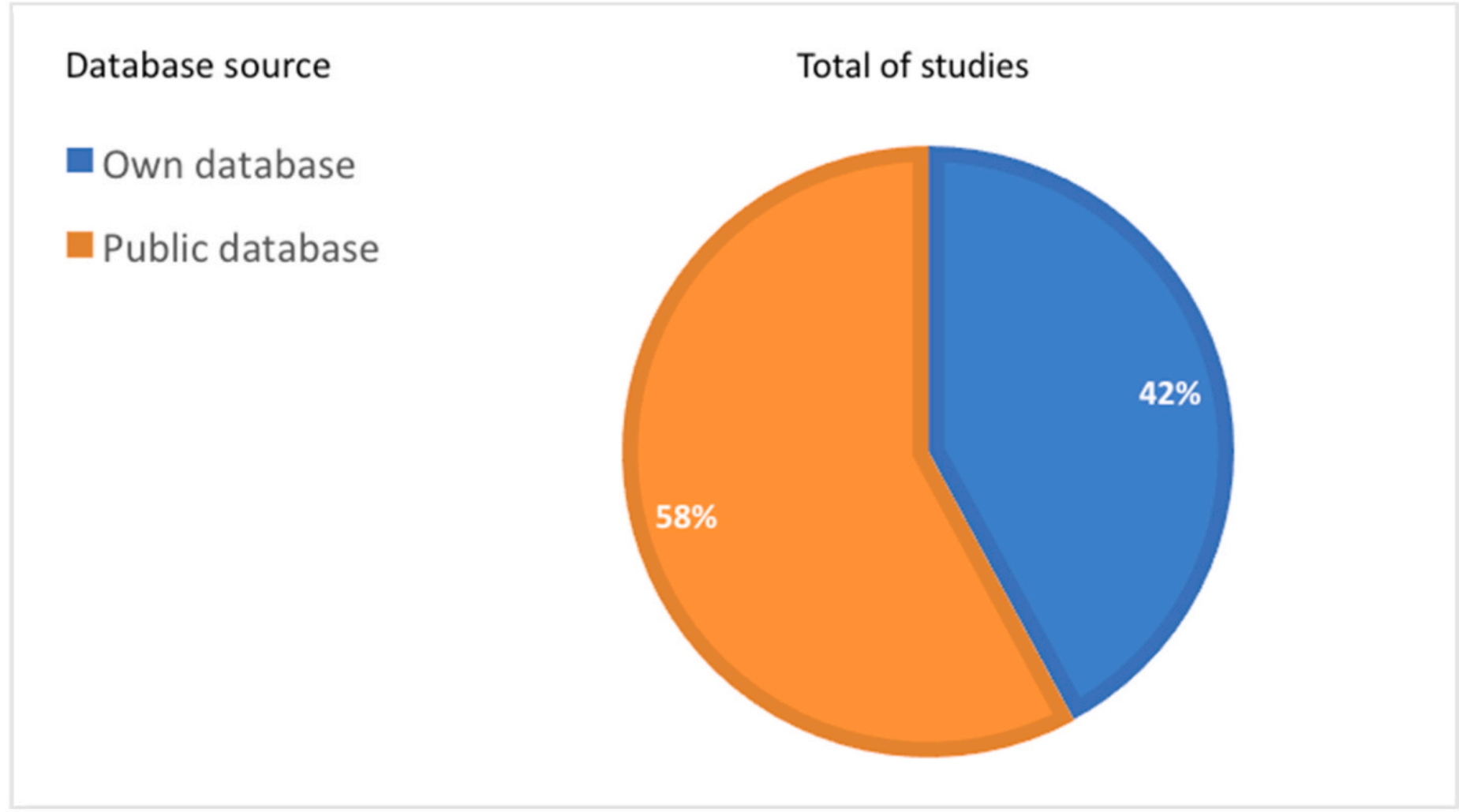

**Fig. 6.** Database creation in relation to RQ7. The research works (%) that developed their own database vs. those that used an existing database are shown.

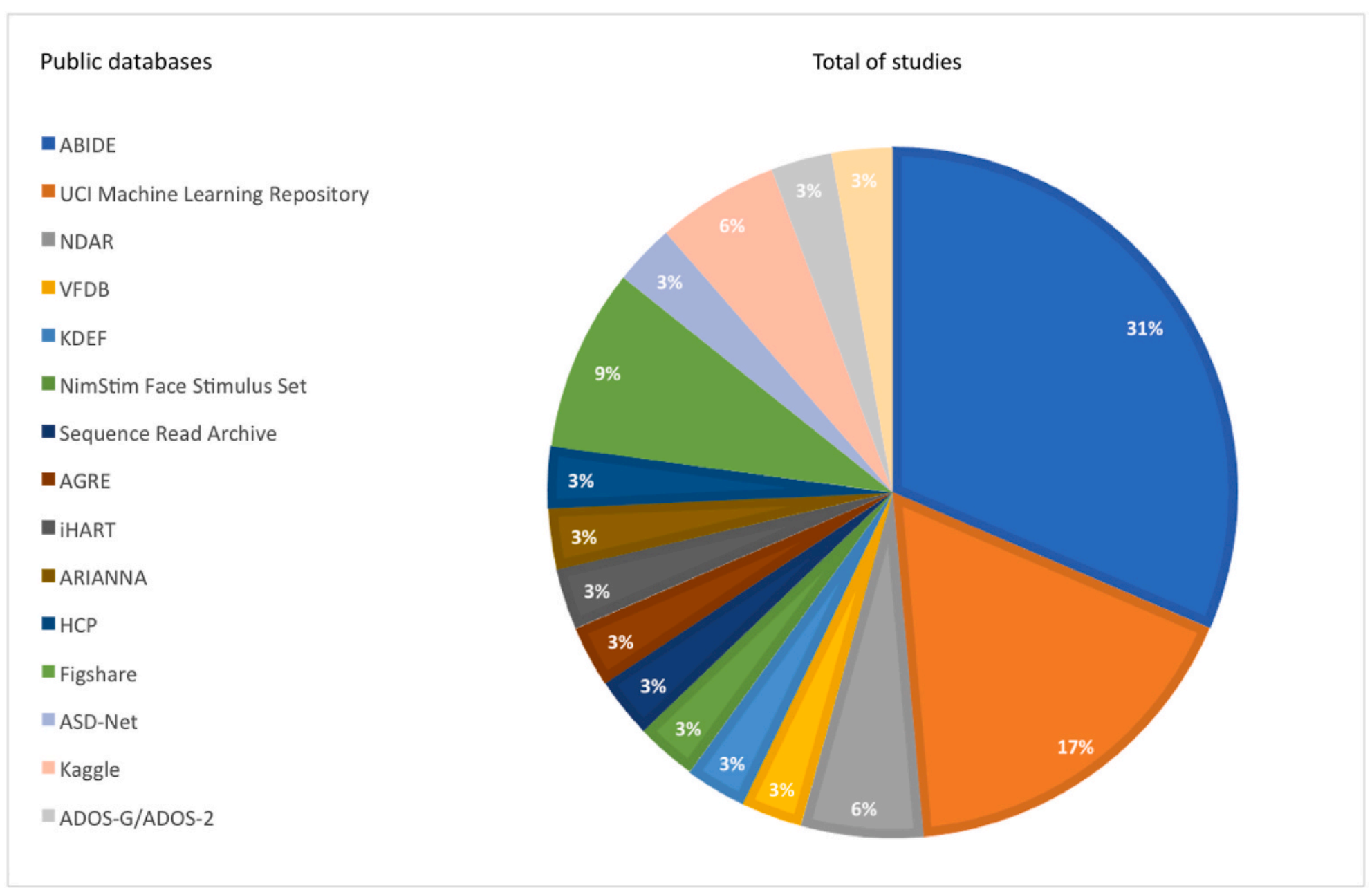

**Fig. 7.** Public databases used in the research studies regarding RQ7.

## 4. Discussion

The application of ML techniques to various research problems improves system performance in different areas. Regarding the ASD problem, research on the use of ML techniques presents challenges due to the large number of variables related to the computational process, which include the specific subdomain corpus, feature selection, learning task, and combination of ML techniques. An appropriate choice of these variables implies the best system performance in this ASD framework, where early diagnosis and treatment by doctors supported by these ML systems will condition the patient's quality of life. Therefore, this study aims to synthesise the literature review process of ML techniques that have been applied to the ASD area during the last seven years (between 2017 and 2023). A total of 55 studies using different ML techniques were identified and reviewed. In the following subsections, we summarise all the RQs analysed and link them to the metrics, challenges and limitations of the proposals studied.

**Table 9**
Public databases used in the research studies related to RQ7.

| Database | Studies |
|---|---|
| ABIDE | [24,26,27,31,34,35,39,44,45,54,56] |
| UCI Machine Learning Repository | [10,28,48,62,67,72] |
| NDAR | [29,55] |
| VFDB | [74] |
| Sequence Read Archive | [40] |
| ARIANNA | [56] |
| KDEF | [51] |
| AGRE | [73] |
| HCP | [45] |
| NimStim Face Stimulus Set | [23] |
| iHART | [73] |
| Figshare | [36,68,70] |
| ASD-Net | [69] |
| Kaggle | [48,71] |
| ADOS-G/ADOS-2 | [59] |

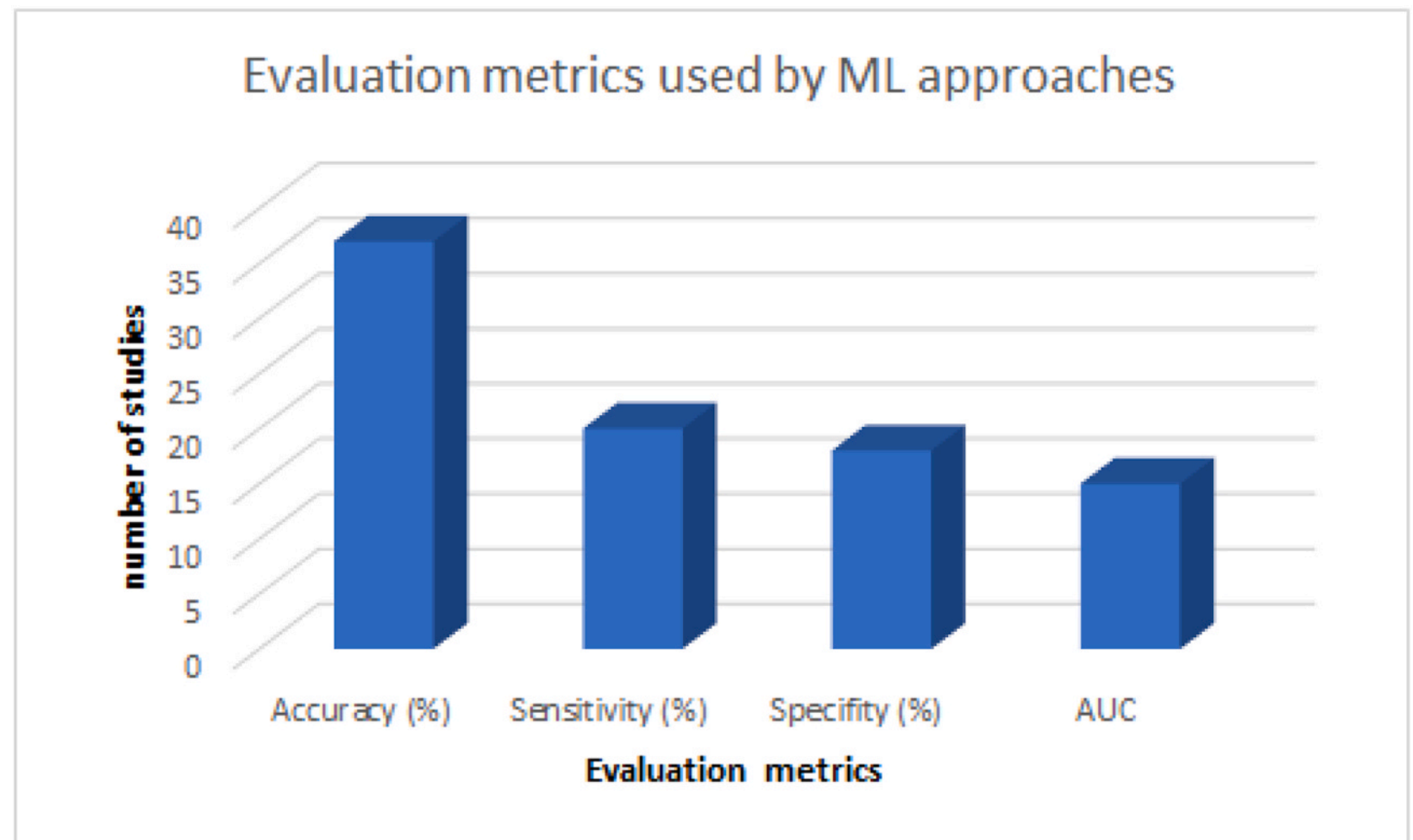

**Fig. 8.** Evaluation metrics used in the research studies with respect to RQ8.

### *4.1. Metrics (RQ1 and RQ2)*

In these studies, journals from the Netherlands published the largest number of research works, followed by the USA and the UK (RQ1). Regarding the countries to which the contributing authors of the studies belonged (RQ2), the USA led this score, closely followed by China, whereas other countries were far behind. These results are unsurprising because of the considerable annual investment in Research and Development[2] of both countries, which is significantly greater than that of other countries were studies in this field were conducted. Consequently, the large number of high-quality researchers from both countries is justified.

### *4.2. Challenges (RQ3, RQ4, and RQ7)*

As discussed in the previous RQ3 section, the supervised approach was the most frequently used ML approach. This outcome was expected, as in many cases, the goal was to diagnose autism. Therefore, in RQ4, the most commonly used ML techniques were those that employed supervised learning methods, such as SVM, RF, RL, ANN, and DT. The recent deployment of techniques based on new technological developments that extend from deep learning should also be highlighted. The latest convolutional network algorithms in their different architectures are expected to provide tools for one of the future challenges in which heterogeneous variables can be combined, from commonly measured ones found in databases (as discussed in RQ7) to new biomarker data trends. As described in this paper, deep learning is one of the most recent areas of research and the quantification of uncertainly remains a major problem in this field. Thus, various studies were conducted in recent years [78], where a novel approach known as the automated random deactivating connective weights approach (ARDCW) was presented. The study was validated using contour maps of the predicted error for different dropouts, accuracy metrics, and success rates, and by comparing with Monte Carlo dropout and quantile regression.

[2] https://asia.nikkei.com/Business/Science/China-passes-US-as-world-s-top-researcher-showing-its-R-D-might (visited on April 16, 2024).

As previously mentioned, the cross-cutting of applicable artificial intelligence methods, leading to the analysis of DNA biomarkers for predicting autism, is worth highlighting [33]. The challenges of data science, as well as the opportunities it offers in strategic partnerships, lead to work such as reported in Ref. [27], where crowdsourcing was leveraged by organising a Kaggle competition to build a pool of ML pipelines for diagnosing neurological disorders. Furthermore, it has been applied to the diagnosis of ASD using cortical morphological networks derived from T1-weighted magnetic resonance imaging. In Ref. [28], the authors used the MGOA Algorithm which can detect ASD in all age groups. Although it is not one of the most common algorithms, it has the potential to explore and exploit the search space effectively. In the work of [79], authors identify clinician subjectivity and lack of data-driven decision-making as key barriers affecting treatment quality. To address these, the application of two ML algorithms is proposed to recommend and personalize treatment goals Applied behaviour analysis (ABA) for 29 study participants with ASD. The authors of [80] present a novel strategy for the early diagnosis of ASD through a prediction model combining RF-CART and RFID3. The test results are very promising as the proposed model outperforms state-of-the-art methods both in the AQ10 dataset and in real-world applications regarding reliability, sensitivity, susceptibility, clarity, and false positive rate. All the challenges discussed in this section will be addressed in future research.

With regard to the types of data utilized in state-of-the-art studies, structured data is typically employed. However, the challenges and opportunities presented by unstructured data are undeniable. Therapeutic records often provide unstructured information, particularly in the form of free-text clinical notes, which contain significant information not captured in structured records. A similar approach is illustrated in the study by Peng et al. [81], where Natural Language Processing (NLP) tools are evaluated using ASD as a case study. CLAMP, cTAKES, and MetaMap are employed to analyse 544 full-text articles and 20,408 PubMed abstracts to extract ASD-related terms. The analysis protocols utilized in this study have broader applicability and can be extended to other neuropsychiatric or neurodevelopmental disorders lacking well-defined terminology sets to describe their phenotypic presentations.

It is crucial to emphasize that the development of efficient and effective ML models for ASD involves close collaboration among computer scientists, clinicians, psychologists, and other stakeholders. Computer scientists contribute their expertise by developing robust algorithms and computational techniques for analysing large datasets related to ASD, while clinicians provide valuable insights into the clinical manifestations of the disorder and the specific challenges faced by individuals with ASD and their families. Psychologists contribute their understanding of cognitive and behavioural aspects of ASD, helping to guide the selection of relevant features and outcome measures for the ML models. Additionally, input from other stakeholders, such as educators, caregivers, and individuals with ASD themselves, ensures that the models are designed with a holistic understanding of the needs and perspectives of the ASD community. By leveraging the diverse expertise and perspectives of these stakeholders, the primary objective is to develop ML models that are not only technically robust but also clinically meaningful and relevant to the real-world challenges faced by individuals with ASD.

In terms of privacy and security considerations, it is important to acknowledge that studies of this nature involve handling sensitive data. It is essential to ensure the protection of participants in medical research, particularly when dealing with individuals with neurodevelopmental disorders. Data collection in this context should be conducted ethically and respecting the privacy of individuals. Informed consent must be obtained from participants or, when applicable, from family members, and the research should be approved by ethical committees of institutions/universities. Data should be stored and processed securely to protect the confidentiality of sensitive medical and behavioural information. It is essential to implement robust security measures to prevent unauthorized access and to maintain data integrity and confidentiality throughout the research process. Additionally, transparency in the use of ML algorithms must be considered, and results should be interpreted and communicated responsibly and ethically, taking into account the potential implications for the health and well-being of individuals with ASD and their families.

In summary, the major challenges of ML-based interventions for ASD include: (a) Generalisability and robustness: ML models can struggle to generalize learned patterns to new settings or populations. This is especially relevant in the case of ASD, where variability in clinical presentation and individual characteristics can be considerable. Models must be able to adapt to this variability to be effective in different clinical contexts; (b) Interpretability and explainability: Often, ML models, especially more complex ones such as deep neural networks, can be difficult to interpret, making it difficult to understand how they arrive at their decisions. This lack of explainability can lead to mistrust and limit the adoption of these tools in clinical practice; (c) Privacy and data security: The collection and processing of sensitive ASD patient data raise ethical and legal concerns related to privacy and data security. It is essential to ensure that patient data are protected from unauthorized access and that privacy standards set by regulations and ethical norms are upheld; (d) Equity and algorithmic bias: There is a risk that ML models reproduce and amplify existing biases in training data, which could lead to unfair disparities in care and treatment for people with ASD. It is important to address these biases through algorithmic bias mitigation and evaluation techniques to ensure equity in access to ML-based interventions; (e) Integration with clinical practice: For ML-based interventions to be effective in clinical settings, they must be seamlessly integrated into existing workflows and accepted by healthcare professionals and patients. Achieving this may require specialized training, updates to technological infrastructure, and careful consideration of the human and social aspects of implementation.

### *4.3. Limitations (RQ5, RQ6, and RQ8)*

Regarding the type of data (RQ5), most studies that applied ML techniques to the problem of autism prediction used brain data, such as those extracted from electroencephalograms. Consequently, the data extraction can sometimes be very challenging, which justifies the use of public databases (58 % overall). ADIBE is one of the most common databases that uses brain data, and it was used by 38 % of the studies that used public databases. Studies that create their own databases tend to include fewer patients because it is sometimes difficult to obtain the opportunity and permission from the ethics committee to work with a larger number of patients (RQ6). Most of the studies considered between 0 and 50 people and between 51 and 100 people (21 % for both cases). To date, structured data are the most frequently used. As suggested in Ref. [82], it would be interesting to consider the possibility of using less structured data with context-sensitive information in the assessment and intervention tasks.

Therefore, the lack of a larger number of databases available to researchers when reproducing their experiments must be noted. One promising trend that will likely take some time to develop is the application of digital twins in healthcare and their generation of synthetic data, which would greatly benefit artificial intelligence [83]. Moreover, in many cases, the genetic basis of ASD must be further explored. However, this task is not straightforward and is challenging as well. Despite the strong genetic basis of ASD, the complete complement of ASD-associated genes is still lacking [84]. In this context, ML and the advances in deep learning may allow researchers to shed some light on this computational complexity.

Finally, regarding the performance of ML techniques in the ASD field (RQ8), various ML techniques have been applied over the years to address challenges in the medical and biomedical domains. This indicates that ML algorithms have been improved, and training corpora have also been enriched qualitatively and quantitatively. Therefore, the continuous number of research works that apply ML techniques in these fields enhances the knowledge of new researchers who use this feedback to improve their new ML techniques. Thus, ML techniques relevant to ASD continue to undergo continuous improvement and evolution.

### *4.4. Innovative approaches*

To conclude this section, we summarise the limitations of ML models applied in ASD:

- Data scarcity: Lack of specific, quality data affects the ability of models to learn representative patterns. This is a constant in the area of health and in particular in the area of autism.
- Data Bias: Biases arise from collecting data from homogeneous populations and this leads to biased models. For example, there are many more autism studies on males than females and this biases the data set and its models.
- Generalisation of models: For these reasons it is very important to always look for the highest generalisability, otherwise models, when tested outside their training samples, will not adapt well to new populations.

In light of these limitations, some possible innovative solutions would include:

- Federated Learning: can present a solution for sharing knowledge without compromising the privacy of patient data, allowing for more robust and general models.
- Development of benchmarking datasets: Suggests collaborative initiatives to create representative datasets, which include a diversity of population samples and ensure that the model is applicable to diverse autism spectrum characteristics.
- Fine tuning and cross-validation techniques: Highlights the use of techniques that can help models fit new data, helping to improve generalisability.

## 5. Conclusion

In this paper, a systematic review of research conducted between 2017 and 2023 was presented, focusing on the application of ML techniques to ASD. The studies reviewed show promising developments, particularly in supervised methods; however, the potential for hybrid methods that integrate unsupervised learning, deep learning, and fuzzy logic is significant and represents an exciting avenue for future research. Hybrid methods could make it possible to discover new patterns and clusters within ASD, an important advancement given the complexity and variability of the spectrum. Rather than simply identifying the presence of autism, these approaches could help define diverse ASD subgroups and degrees of dependency, supporting a more nuanced approach to diagnosis.

To ensure hybrid methods become more accessible and interpretable, future research will need to address the integration of diverse data sources. This will require attention to three key areas: (1) the use of less structured data that includes context-sensitive information, (2) digital twins and synthetic data generation, and (3) genetic data analysis combined with deep learning to uncover new features and unknown correlations. Future ML approaches should also incorporate new data collection methods, including wearable technologies (such as biometric sensors) that enable continuous and non-intrusive monitoring of behaviour and physiological responses in natural environments. Additionally, integrating multimodal data —clinical records, genetic data, medical imaging, and behavioural records— will enable a holistic understanding of ASD. However, this integration introduces technical and ethical challenges that must be addressed.

Further exploration is necessary to understand how evolving ML techniques can meet specific challenges in ASD diagnosis and treatment. This includes advancements in deep learning, natural language processing for analysing textual data, and semi-supervised or reinforcement learning methods adapted to clinical datasets. The potential for interpretable and explainable ML models is particularly crucial in autism research, as these could improve confidence among clinicians and patients in the decision-making processes supported by these tools. Specifically, in the last two years (2024 and 2025), the application of ML and deep learning techniques to ASD detection and assessment has become a consolidated trend [85–90]. Some studies explore models with a reduced number of features to improve efficiency, while others rely on neural networks and bidirectional long short-term memory architectures for classification. Additional approaches incorporate functional magnetic resonance imaging data or eye-tracking information to enhance diagnostic performance. Considering the most common evaluation metrics (accuracy, F1-score, sensitivity and specificity), these recent works consistently demonstrate that the use of ML and deep learning for ASD diagnosis remains a robust and promising research direction.

Finally, while ML offers opportunities to improve ASD detection and diagnosis, applications extend beyond these areas. Emerging ML applications in public health, treatment support, and clinical research show strong potential, as initial results indicate. The ongoing expansion of ML tools requires multidisciplinary teams —including experts in computer science, psychology, neuroscience, and medicine— to collaboratively tackle the complexities of ASD and to develop customized, context-sensitive approaches for personalised diagnoses, treatments, and interventions. The insights presented in this study establish a foundation for future research and progress in this growing field, providing a roadmap for advancing ASD research through machine learning.

## CRediT authorship contribution statement

**Rafael Muñoz-Terol:** Writing – review & editing, Writing – original draft, Investigation, Funding acquisition. **Jesús Peral:** Writing – review & editing, Writing – original draft, Investigation, Funding acquisition. **Sandra Amador:** Writing – review & editing, Writing – original draft, Investigation. **David Gil:** Writing – review & editing, Writing – original draft, Investigation, Funding acquisition.

## Ethics declaration

Review and/or approval by an ethics committee as well as informed consent was not required for this study because this literature review only used existing data from published studies and did not involve any direct experimentation/studies on living beings.

## Data availability statement

No data was used for the research described in the article.

## Funding

This research has been funded by BALLADEER Project (PROMETEO/2021/088) from the Conselleria de Innovación, Universidades, Ciencia y Sociedad Digital, Generalitat Valenciana (Valencia, Spain). Furthermore, it has been supported by the KOSMOS-UA project (PID2024-155363OB-C43), funded by the Spanish Ministry of Science and Innovation, the BALIDA-AA project (CIPROM/2024/13), funded by Conselleria de Educación, Cultura, Universidades y Empleo (Generalitat Valenciana), the IAEAV project (INREIA/2024/176), funded by the Conselleria de Innovación, Industria, Comercio y Turismo (Generalitat Valenciana), the ENIA Chair of Artificial Intelligence (TSI-100927-2023-6), the AgroVAL (TSI-100122-2024-10), the Sophia (TSI-100130-2024-10) and European mobility for efficient planning and new business opportunities (TSI-100121-2024-10) projects, funded by the Recovery, Transformation and Resilience Plan from the European Union Next Generation through the Ministry for Digital Transformation and the Civil Service, and Grant RED2022-134656-T, funded by MCIN/AEI/10.13039/501100011033.

## Declaration of competing interest

The authors declare that they have no known competing financial interests or personal relationships that could have appeared to influence the work reported in this paper.